\documentclass[letterpaper, 10 pt, conference]{ieeeconf}  

\IEEEoverridecommandlockouts                              
\usepackage{graphicx} 
\usepackage{epsfig} 
\usepackage{times} 
\usepackage{amsmath} 
\usepackage{amssymb}  
\usepackage{makecell}
\usepackage{mathtools}
\usepackage[symbol]{footmisc}

\DeclareMathSizes{10}{10}{7.5}{7}
\usepackage{svg}

\usepackage{dsfont}
\usepackage{algorithm}
\usepackage[noend]{algpseudocode}
\usepackage{hyperref}
\usepackage{csquotes}

\usepackage{stmaryrd}
\usepackage{trimclip}

\title{\LARGE \bf
Augmenting Imitation Experience via Equivariant Representations
}

\author{
Dhruv Sharma$^{*\dag}$
\quad
Alihusein Kuwajerwala$^{*\dag}$
\quad 
Florian Shkurti$^{\dag}$
\thanks{$^*$Authors contributed equally.}
\thanks{$^{\dag}$Robot Vision and Learning Laboratory, University of Toronto Robotics Institute. 
{\tt\small \{dhruv, florian\}@cs.toronto.edu, ali.kuwajerwala@mail.utoronto.ca}}
}

\begin{document}

\maketitle
\thispagestyle{empty}
\pagestyle{empty}

\begin{abstract} 
The robustness of visual navigation policies trained through imitation often hinges on the augmentation of the training image-action pairs. Traditionally, this has been done by collecting data from multiple cameras, by using standard data augmentations from computer vision, such as adding random noise to each image, or by synthesizing training images. In this paper we show that there is another practical alternative for data augmentation for visual navigation based on extrapolating viewpoint embeddings and actions nearby the ones observed in the training data. Our method makes use of the geometry of the visual navigation problem in 2D and 3D and relies on policies that are functions of \emph{equivariant} embeddings, as opposed to images. Given an image-action pair from a training navigation dataset, our neural network model predicts the latent representations of images at nearby viewpoints, using the equivariance property, and augments the dataset. We then train a policy on the augmented dataset.  Our simulation results indicate that policies trained in this way exhibit reduced cross-track error, and require fewer interventions compared to policies trained using standard augmentation methods. We also show similar results in autonomous visual navigation by a real ground robot along a path of over 500m. 

\end{abstract}

\section{INTRODUCTION}

\begin{figure}[t]
\centering
\includegraphics[width=0.95\columnwidth]{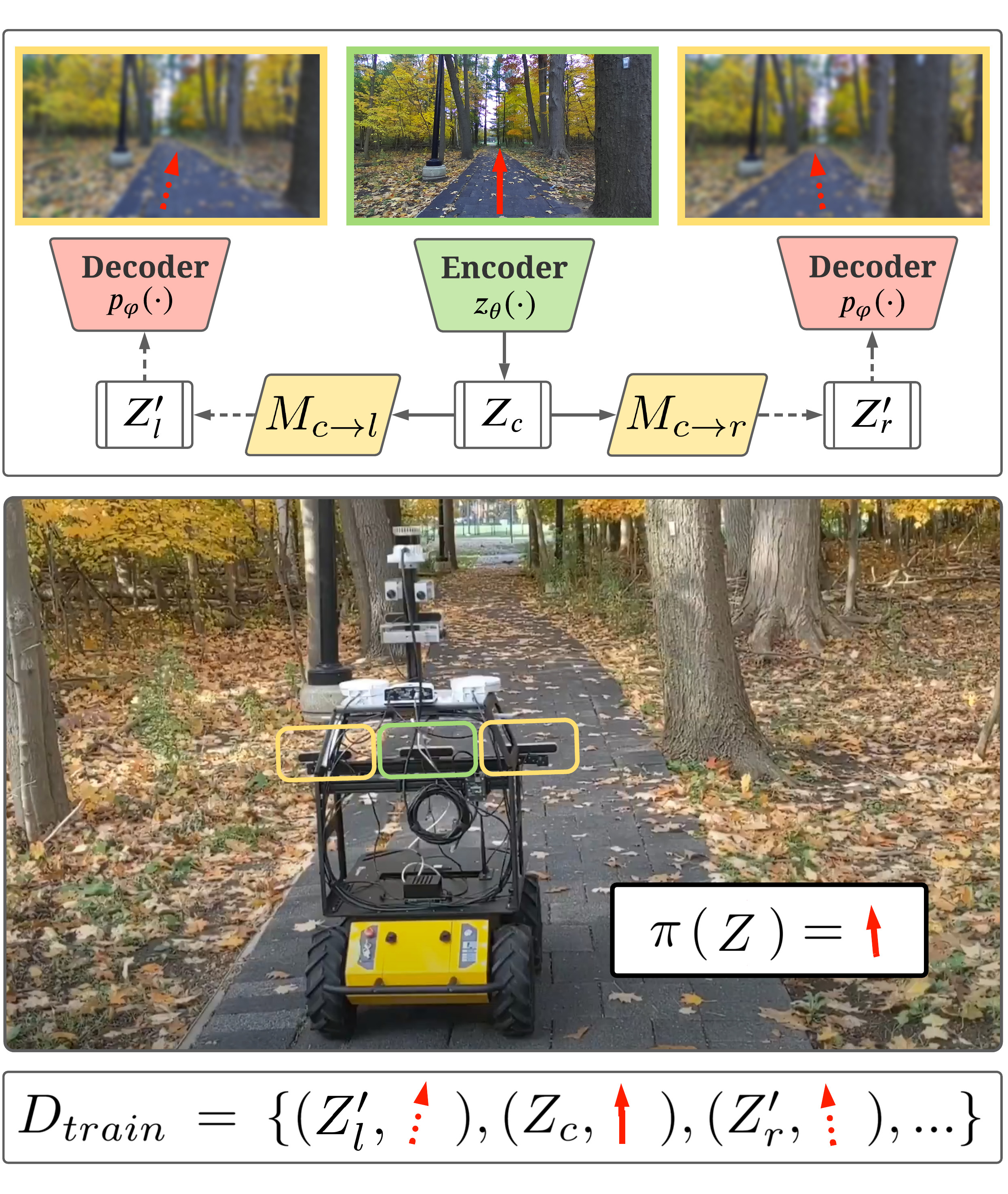}
\vspace{-0.3cm}
\caption{An overview of our system. Images from the center camera are embedded into a low-dimensional representation, $Z_c$, which is deterministically transformed into the predicted embeddings of the left and right viewpoints respectively, via equivariant mappings $M$. Both the actual steering command (solid red arrow) and the predicted steering commands (dashed red arrows) are included in the augmented training set on which the policy is trained. At test time, the policy is executed on embeddings from the center camera. The maps $M$ are trained separately from $\pi$.}
\label{fig:first_page}
\vspace{-0.5cm}
\end{figure}

From autonomous driving to flying and manipulation, vision-based policies learned through imitation are starting to be deployed on robot systems of critical importance. To ensure safety during deployment, we would ideally like to have formal guarantees about the generalization and robustness properties of these policies. While desirable, however, formal guarantees for out-of-distribution generalization have been attained on perturbation models that are theoretically convenient, but limited in practice~\cite{pmlr-v97-cohen19c}, as they often do not capture the full range of plausible unseen states. Data augmentation during training is significant in practice. 

In this paper we address this problem by proposing a dataset augmentation scheme for training policies whose inputs are image embeddings, as shown in Fig.~\ref{fig:first_page}. 
Given an observed image from the center camera of a vehicle, our method allows us to learn a map that predicts the embeddings of corresponding nearby viewpoints, as well as the actions that would have needed to be taken if the vehicle's camera was at those viewpoints. Our method relies on predicting embeddings of nearby viewpoints using \emph{equivariant mappings}~\cite{Jayaraman_2015_ICCV, lenc2015understanding} that are trained to transform the embedding of the center camera to the embeddings of nearby viewpoints, both for ground vehicles and for flying vehicles, as we will show in the evaluation section. Equivariance has emerged as an important structural bias for neural network design in the last few years. Here we use it as an auxiliary loss to the main behavioral cloning loss. 

Our main contribution is showing that augmenting the training dataset with predicted embeddings of nearby viewpoints, using learned equivariant maps, increases the robustness of the vision-based policy. This results in lower cross track error and fewer human interventions needed for navigation tasks, both for ground vehicle and aerial vehicle navigation tasks in 2D and 3D respectively. We demonstrate that our method applies to real world visual navigation scenarios by deploying it on a terrestrial mobile robot, resulting in significantly improved navigation performance.

\section{RELATED WORK}
\textbf{Equivariant representations:} Equivariance for image representations captures relationships between encodings of two related images that are determined by the transformation of those images~\cite{lenc2015understanding}. If the encodings are the same, then they are invariant to that transformation. In our work, we want the representation of the left camera to be predictable by the center camera, so we want a high degree of equivariance. Feature equivariance and invariance in terms of rotation transformations was studied in~\cite{equivariant_rotations, Zhao2019QuaternionEC, lifd_survey}, as well as in the context of probabilistic models~\cite{dbn_equivariant, Bilovs2020EquivariantNF, Rezende2019EquivariantHF}. The utility of equivariant representations for multiple downstream tasks, learned by egomotion was recognized in~\cite{Jayaraman_2015_ICCV}. Our work here is related the most to this paper.

Equivariant representations have also been useful for 3D volumetric data~\cite{Weiler20183DSC}, even in the fully unsupervised setting~\cite{Spezialetti2019LearningAE}, addressing equivariance and invariance to both rotation and translation for 3D data~\cite{Fuchs2020SE3Transformers3R, Wang2020EquivariantMF}.
Additionally, instead of imposing equivariance through a loss function, as in~\cite{Jayaraman_2015_ICCV}, many recent works build equivariance in the structure of the network~\cite{cubenet, worrall2017harmonic, cohen2019gauge, Tai2019EquivariantTN, Cohen2019AGT, steerable_filters_equivariant_cnn}, for example in the multi-view setting~\cite{emvn, multiviews, gvcnn, qi2016volumetric, you2018pvnet}. Finally, equivariance has also been used to improve state representations for planning in MDPs~\cite{Pol2020PlannableAT}.

\textbf{Automated data augmentations:} Typical dataset augmentations for supervised deep learning have included addition of noise, rotations, crops, blurring, and others. While this is usually a set of manually hand-crafted augmentation schemes, there has been increased interest in automatically computing data augmentations~\cite{ratner2017learning, zhang2019adversarial, dao2019kernel, raileanu2020automatic} to increase the robustness of classifiers, regressors, and RL policies. Augmentations have also been used in monocular depth estimation~\cite{ravi_unsupervised_depth}. 



\textbf{Robustness in behavioral cloning:}
Behavioral cloning techniques that learn policies through pure supervised learning from a given dataset of state-action pairs usually suffer from covariate shift~\cite{Loquercio_2018,lecun_obstacle_avoidance}. Dean Pomerleau's PhD thesis~\cite{pomerleau_thesis} and pioneering work on ALVINN~\cite{alvinn} details a data augmentation scheme that injects synthetic images of the road ahead from viewpoints nearby the camera to the training set. Our paper provides an alternative that is based on neural network view synthesis, which in the last years has seen significant progress. Pomerleau's thesis also identified issues of covariate shift, which methods such as DAgger~\cite{DAgger_ross} address by iteratively labeling the newly visited states after each policy evaluation. The question of dataset augmentation in one round of behavioral cloning, however, remains. Collecting large driving datasets, in a way that no manual annotation is required from the user, is critical for autonomous visual navigation. 

\begin{figure*}[t]
\centering
\includegraphics[width=\textwidth]{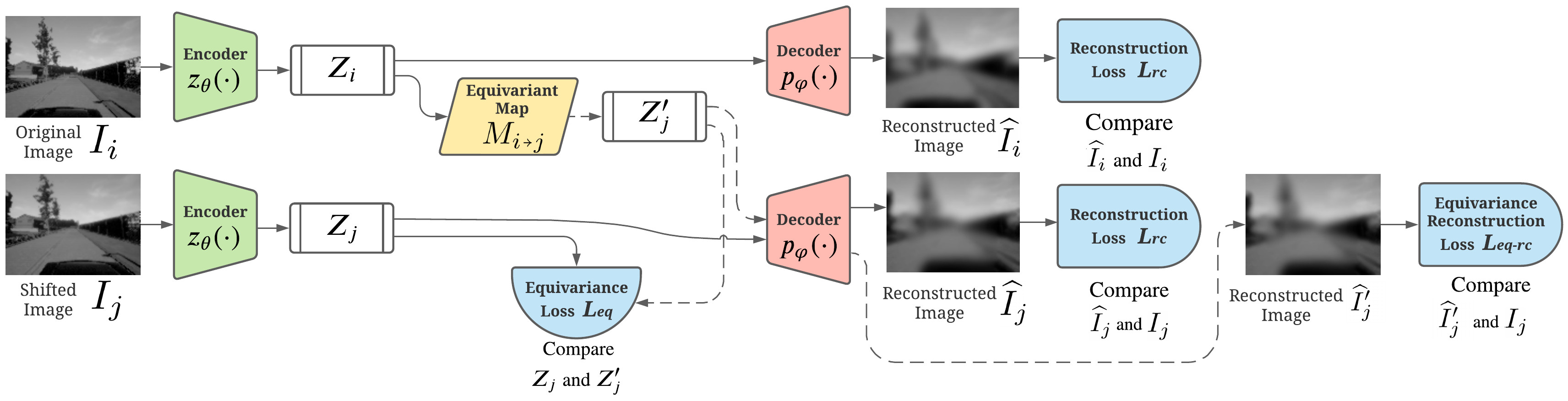}

\vspace{-0.4cm}

\caption{Representation Learning Phase: we use an encoder-decoder network to learn image embeddings that are equivariant to input camera viewpoint. Mapping network $M_{i \rightarrow j}$ is learned to map these embeddings from viewpoint $i$ to viewpoint $j$. The encoder-decoder and the mapping networks are trained collectively by optimizing a combined loss function shown in Eqn. \ref{eqn:repr_loss}. See~\ref{methods} for details.}
\label{fig:eqSetup}
\end{figure*}

One example of this was in UAV navigation in forest trails~\cite{trail_following}. The images of the center camera are automatically annotated with the forward steering command, the images of the left camera with the right, and the images of the right camera are annotated with the left steering command. A similar hardware setup and automated data collection was used by the NVIDIA self-driving car team~\cite{nvidia_driving}, as well as more recent work on behavior cloning conditioned on human commands (e.g. turn left at the next intersection)~\cite{codevilla_2017}. 
The crux behind the idea of the 3-camera setup is that behavioral cloning can become more robust if we obtain expert actions from states that are nearby the forward-facing one.
Our method builds upon this approach by replacing the need for additional sensors with a model that predicts their embeddings, whose corresponding actions are automatically annotated. We note that in other similar approaches, such as DART~\cite{dart_laskey} an expert demonstrator will have to manually annotate the additional states. In DART, Gaussian noise is added to the original training dataset in order to explicitly include nearby labeled states in the augmented dataset, together with manual annotations.

\section{METHODOLOGY} \label{methods}
Our goal is to learn an encoding method that satisfies equivariance constraints with respect to camera poses or viewpoints nearby the original camera. This allows us to predict the embeddings of nearby viewpoints from the observations of the center camera.
Here, nearby viewpoints refer to horizontal and vertical translation of the camera over short distances capturing the same scene. Our case involves translations of 0.25 to 0.5 meters.

Given an image $I_i$ of a scene taken from viewpoint $i$, we compute its corresponding embedding in the latent space $Z_i$, and use an equivariant map $M_{i \rightarrow j}$ to transform it to the embedding corresponding to 
another desired nearby viewpoint $j$. This predicted embedding $Z\sp{\prime}_j$ can then be used as 
augmented training data for policies trained on such embeddings and actions.


We train an image encoder network on a dataset of image pairs, where each pair consists of images of  the same scene from neighboring viewpoints. In our case, the images are obtained from synchronized videos recorded by a set of cameras mounted on a vehicle visually navigating through an environment. The relative poses of the cameras are assumed to be known and looking at the same scene. With each image pair we simultaneously also record the corresponding expert control command used to navigate the vehicle at that time.

We then use the learned embeddings to train policies using imitation learning.

\subsection{Equivariant Feature Learning} \label{eqLearning}

Given a tuple of images $\textbf{I} = \langle I_1, I_2, ..., I_n \rangle$ that capture the scene at the same timestep, our model learns an encoder $z_\theta (\cdot): I_i \to \mathbb{R}^D$, parametrized by $\theta$, that maps an image to an embedding $Z_i$. We want the embeddings to exhibit equivariance, i.e. change predictably with respect to transformations applied to the viewpoint $i$ of the input image $I_i$. Let $g$ be a transformation applied to the input image $I$ in the pixel space, and let $M_g$ define a corresponding transformation in the latent space. Then:
\begin{equation}
    z_\theta (g(I)) \approx M_g (z_\theta(I))
\end{equation}
For example, suppose the images $I_c$ and $I_l$ correspond to the center and left camera images in Fig.~\ref{fig:first_page}, respectively. If $g$ is an image transformation such that \mbox{$g(I_c) = I_l$}, then $M_{g}=M_{c \rightarrow l}$ defines a transformation in the latent space such that $ M_{c \rightarrow l}(Z_c) = Z\sp{\prime}_{{ l}} \approx Z_l$. In latent space, we want $Z_c$ and $Z_l$ to maintain a geometric relation directly corresponding to the one between images $I_c$ and $I_l$.



\begin{figure}[t]
\centering
\includegraphics[width=\columnwidth]{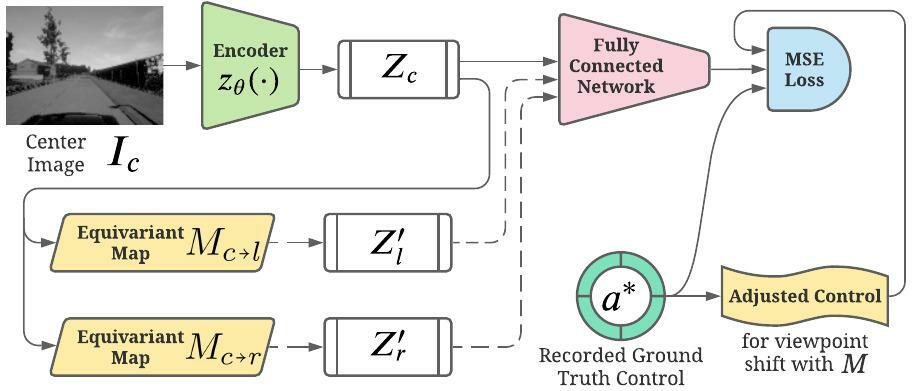}

\vspace{-0.3cm}


\caption{Policy Training Phase: The center image is passed to an encoder, then the resulting embedding is mapped to nearby viewpoints (for experiments using an equivariant map), and finally the embeddings are passed to a FCN model to produce steering output. Note that the encoder and FCN models are trained \textit{separately}. See~\ref{bcTraining} for details.}
\label{fig:controlsetup}
\vspace{-0.5cm}
\end{figure}

\subsection{Equivariant Map Learning as an Auxiliary Task} \label{eqObjective}

Now, we design an objective function that encourages the learned embeddings to exhibit equivariance with respect to camera transformations, or camera viewpoints in our case. Given a tuple of images $\textbf{I} = \langle I_1, I_2, ..., I_n\rangle$ captured at the same time, where image $I_i$ is taken from the $i^{th}$ viewpoint, and the corresponding embeddings $z_{\theta}(I_i)$, we set up the following objective function to enforce equivariance between any two image viewpoints.
\begin{equation}
    L_{eq}(\theta, M) = \textstyle \sum\limits_{i} \sum\limits_{j} ||z_{\theta}(I_j) -  M_{i \rightarrow j}(z_{\theta}(I_i))||^2
    \label{eqn:leq}
\end{equation}

\subsection{Image Reconstruction as an Auxiliary Task} \label{mappingFun}

We make use of image reconstruction as an auxiliary task, so that the embeddings capture the information needed to reconstruct images across all different viewpoints. We jointly train an encoder-decoder model for each viewpoint, as shown in Fig.~\ref{fig:eqSetup}. The encoder, decoder, as well as the equivariant mapping $M_{i \rightarrow j}$ are optimized together in a shared objective.  
The latent representations $Z_i$ are decoded back to images by a decoder $p_\varphi (\cdot)$. The output of the decoder is then used to compute and minimize the autoencoding loss $L_{rc}$ from the same viewpoint:
\begin{equation}
    L_{rc}(\theta, \varphi) = \textstyle \sum\limits_{i} ||I_i - p_{\varphi}(z_{\theta}(I_i))||^2
    \label{reconLoss}
\end{equation}
\noindent To further reinforce the effectiveness of the equivariant maps $ M_{i \rightarrow j}$, we enforce an additional reconstruction penalty between pairs of viewpoints:
\begin{equation}
    L_{eq\textrm{-}rc}(\theta, \varphi, M) = \textstyle \sum\limits_{i} \sum\limits_{j}  ||I_j - p_{\varphi}(M_{i \rightarrow j}(z_{\theta}(I_i))) ||^2
    \label{eqReconLoss}
\end{equation}
The overall representation learning loss is a combination of the reconstruction, equivariant reconstruction, as well as the equivariant loss:
\begin{equation}
\begin{multlined}
    L(\theta, \varphi, M) = \\ \lambda_1 L_{rc}(\theta, \varphi) + \lambda_2 L_{eq}(\theta, M) + \lambda_3 L_{eq\textrm{-}rc}(\theta, \varphi, M)
    \label{eqn:repr_loss}
\end{multlined}
\end{equation}
\noindent We backprop through the combined loss in Eqn.~\ref{eqn:repr_loss} to optimize $\theta, \varphi$, and $M$. The hyperparameters are chosen as $\lambda_1=1$, $\lambda_2=10$, and $\lambda_3=1$ to ensure all loss terms have relatively same scale.

\subsection{Imitation Learning with learned Features} \label{bcTraining}

After we learn $z_\theta (\cdot)$ using the representation learning loss in Eqn.~\ref{eqn:repr_loss}, we fix the encoder weights and learn a control policy, i.e. a multi-layer perceptron representing $\pi (Z)$, via imitation learning. The control policy $\pi (\cdot)$ is trained on embedding-action pairs both from the center camera (observed embeddings and actions) and from the nearby viewpoints (predicted embeddings from equivariance relations and predicted actions). The objective function used is the following:
\begin{equation}
    L(\pi) = \textstyle \sum\limits_{i} ||a_i - \pi(Z_i)||^2
    \label{eqn:policy_loss}
\end{equation}
At test time and deployment, the encoder is used to convert the images to embeddings which are then used as input to the learned policy, as shown in Fig.~\ref{fig:eqSetup}.





\section{Experiments}
We validate our approach in two different experimental settings, an aerial vehicle and a terrestrial vehicle. The experiments involve the following steps: 

\textbf{Data Collection.} Each data sample $\langle \textit{\textbf{I}}, \textit{\textbf{A}} \rangle$ consists of an image tuple $\textit{\textbf{I}}$ that consists of images from different camera viewpoints and an action set $\textit{\textbf{A}}$ that contains the corresponding control actions. We collect data both in simulation and in the real world, for the respective experiments.
    
\textbf{Learning Equivariant Features.} We learn the feature mapping function $z_\theta(\cdot)$ and equivariant mappings $M_{i \rightarrow j}$ using the collected image sets $\textit{\textbf{I}}$ in the original training dataset.

\textbf{Policy Evaluation.} In order to assess the effectiveness of our learned embeddings and equivariant mappings we evaluate and compare three imitation learning policies that each learn to map images to control actions:

\begin{enumerate}
    \item The first policy is trained on image-action pairs obtained from a single camera, in our case the center camera mounted on the vehicle.
    \item The second policy is trained on image-action pairs obtained from all the camera sensors mounted on the vehicle. In our case it is 3 viewpoints for the car and 5 for the flying vehicle. The recorded expert control is modified to account for the shifted camera viewpoints for the off center cameras.
    \item The third policy uses our equivariant augmentations and is trained on image-action pairs obtained from the center camera and predicted nearby embeddings. The center camere embedding is mapped to the embeddings of the other viewpoints in the latent space. This corresponds to the training scheme described in Sec.~\ref{mappingFun}.  
\end{enumerate}

\textbf{Evaluation Metrics.} 
To compare the three policies, we record the number of interventions (which occur when the ground vehicle goes off of the current lane or when the flying vehicle deviates more than 2.5 meters away from the goal trajectory), average cross track error (deviation from the specified trajectory in meters), and autonomy. 
Autonomy here is calculated as in~\cite{nvidia_driving}, i.e. the percentage of time the car was not being driven by the learned policy due to an intervention. More specifically, for an experiment with $n$ interventions: 
\begin{equation}
    \textrm{autonomy} = ( 1 - \frac{ n \cdot (6 \textrm{ seconds  })}{ \textrm{ elapsed time in seconds} } ) \cdot 100 
    \label{eqn:policy_loss2}
\end{equation} 
The experiment setup details and results are discussed in Sec. \ref{droneExp} and Sec. \ref{carExp}. 

\textbf{Policy and Encoder Architectures.} The network architecture used for vehicle navigation was similar to the one used by NVIDIA in \cite{nvidia_driving}, it consists of a CNN encoder network followed by a FCN regressor network. An important distinction between their work and ours is that the entire model in \cite{nvidia_driving} is trained end-to-end, whereas we train the encoder separately from the MLP, as mentioned in Sec~\ref{bcTraining}. For consistent comparisons, we ensure that the encoder $z_{\theta}(I)$ has the same architecture as the image encoder in \cite{nvidia_driving}. 

\textbf{Policy Training.} All the networks are optimized using ADAM optimizer~\cite{adam} with a learning rate of $10^{-4}$ and a batch size of $64$. We use DAgger \cite{DAgger} as our imitation learning algorithm and we investigate the performance of the control policy with respect to DAgger iterations. As the DAgger runs increase, the difference in performance between the policies becomes less pronounced as the data aggregation process itself makes all three policies more robust. This provides us with one measure of how effective our embedding method is for data augmentation, particularly in the beginning of imitation. 



\subsection{Quadrotor Simulation Experiments} \label{droneExp}
\textbf{Experimental Setup.} For this set of experiments, we use a flying vehicle in the AirSim simulator \cite{airsim2017fsr}, and also a more complex Drone-Racing environment developed by Microsoft \cite{droneracing}, consisting of outdoor three dimensional racing trail marked by rectangular gates. The vehicle is set up with five front facing cameras.
For the dataset used in these experiments, each data sample comprises of an image set $\textit{\textbf{I}} = \langle I_l, I_c, I_r, I_t, I_b \rangle$ corresponding to timestamped images from the left, center, right, top, and bottom cameras. 
The action tuple $\textit{\textbf{A}} = \langle \delta \textrm{yaw}, \delta z \rangle$, is the expert control used to navigate the flying vehicle, applied in terms of the relative change in yaw and $z$ at each timestep. The augmented actions for non-center cameras are computed as shown in Table \ref{tab:flying-corrections}. Note that clockwise rotation is positive $\delta yaw$ and downwards direction is positive $\delta z$. 

\begin{table}[!t]
\caption{out-of-distribution experiment results (flying vehicle)}
\begin{center}
\begin{tabular}{|c|c|c|c|}
\hline
\textbf{Training Setup} & \textbf{Interventions}& \textbf{Autonomy(\%)}& \makecell{\textbf{Cross Track} \\ \textbf{Error (m)}} \\
\hline
Center Camera & 3 & 84.23 & 2.12 \\
\hline
\makecell{Equiv. Augment. \\Out-Of-Dist \\ Training} & 2 & 89.46 & 1.39 \\
\hline
\makecell{Equiv. Augment. \\ In-Dist \\ Training} & 2 & 89.51  & 1.26 \\
\hline 
All Cameras & 0 & 100 & 0.17 \\
\hline
\end{tabular}
\label{tab:eqfrozen-flying}
\end{center}
\vspace*{-0.4cm}
\end{table}

\begin{table}[!t]
\caption{Augmentation label calculations (flying vehicle)}
\begin{center}
\vspace*{-0.2cm}

\begin{tabular}{|c|c|c|c|}
\hline
\textbf{Center Camera} & \textbf{\textit{$\delta yaw$ $(rad)$}}& \textbf{\textit{$\delta z$ $(m)$}} \\
\hline
Left Camera & $\delta yaw+0.03$ $rad$ & $\delta z$ \\
\hline
Right Camera & $\delta yaw-0.03$ $rad$ & $\delta z$ \\
\hline
Top Camera & $\delta yaw$ &  $\delta z + 0.5$$m$ \\
\hline 
Bottom Camera & $\delta yaw$ & $\delta z - 0.5m$ \\
\hline
\end{tabular}
\label{tab:flying-corrections}
\end{center}
\vspace*{-0.7cm}
\end{table}

\begin{figure*}[!t]
\hspace{0.8cm}\includegraphics[width=7in, trim={4.5cm 0.0cm 0.0cm 0.5cm},clip]{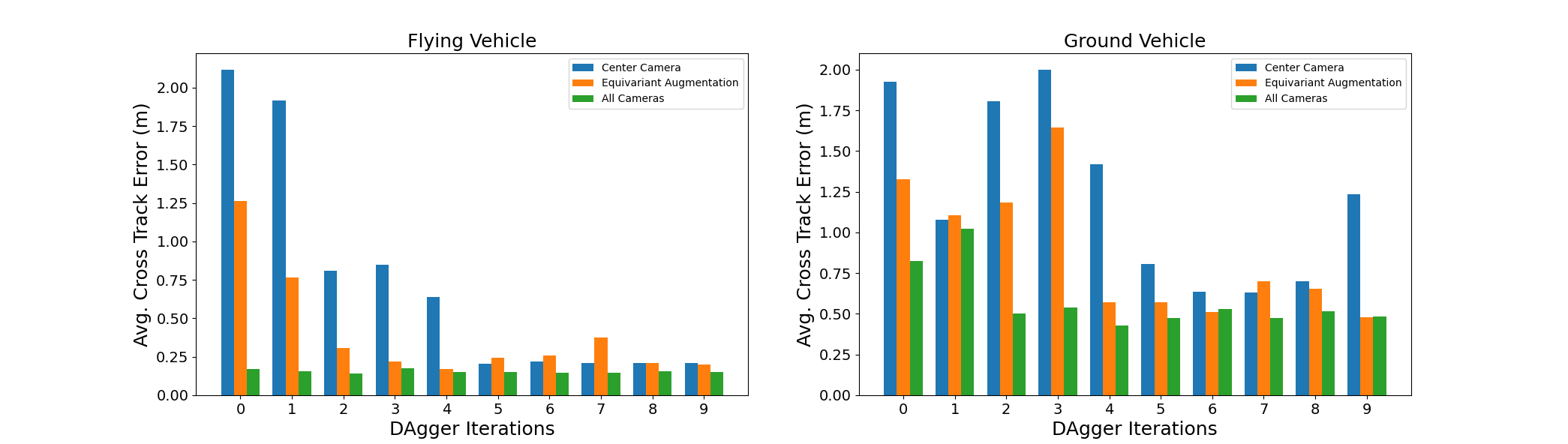}
\vspace{-0.7cm}
\caption{Cross track error (m) vs DAgger iterations for flying vehicle (left) and ground vehicle (right). It can be seen that network trained with equivariant augmentation lies in between the network trained using center camera and all cameras in terms of performance.}
\label{fig:cte_drone}
\end{figure*}

\begin{figure}[!t]
\centering
\includegraphics[width=3.5in]{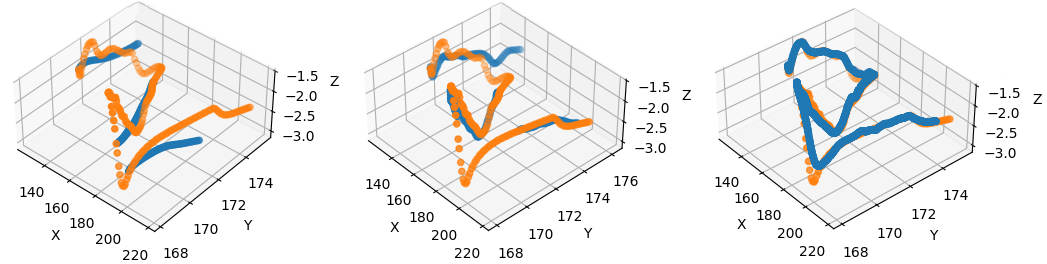}
\caption{Reference trajectory (orange) vs flown trajectory (blue). The all-cameras model (right) tracks the trajectory the best, followed by the equivariant-augmentation model (center), followed by the center-camera model (left). Note that missing sections in the flown trajectory are due to an intervention where expert control was used to navigate the vehicle back to the reference trajectory. See~\ref{droneExp} for details.}
\label{fig:trajs}
\end{figure}

\begin{figure}[!t]
\centering
\includegraphics[width=0.8\columnwidth]{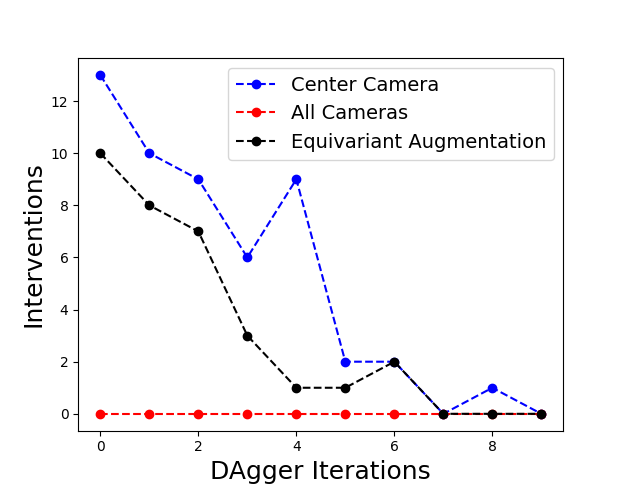}
\caption{Number of interventions vs DAgger iterations for the flying vehicle in drone-racing environment. Model trained using equivariant augmentations lies in between the model trained using a single cameras and the model trained using all five cameras.}
\label{fig:droneracingDAgger}
\end{figure}

\textbf{Results.} Fig. \ref{fig:cte_drone} (left) shows the cross track error vs DAgger iterations for three different test scenarios. We see that using the augmentations generated via the equivariant maps for training improves the imitation performance. Fig. \ref{fig:trajs} overlays the trajectory flown on top of the reference trajectory. The policy trained with equivariant augmentations is able to track better than a policy trained using only the center-camera data. Additionally, as shown in Fig. \ref{fig:droneracingDAgger}, we observe that the policy trained on our augmentations requires on average 2 fewer interventions when compared to the policy trained without them, for the Drone Racing environment.

\textbf{Out-Of-Distribution Experiments.} We perform another set of experiments where we train the equivariant map on a dataset captured in a separate area of the simulation map, a dense urban environment with buildings and roads, compared to where we test the model, a park with no city structures except a single paved path.
As shown in \mbox{Table \ref{tab:eqfrozen-flying}}, the policy trained using the generated augmentations has a lower cross track error by 0.73 meters when compared to the policy trained directly on the center-camera data. 
Furthermore, its cross track error is only 0.13 meters higher than the policy which uses the equivariant map trained on the same area of the simulation map as used for testing. 

\subsection{Driving Simulation Experiments} \label{carExp}


\textbf{Experimental Setup.} For this experiment, we use the Carla simulator \cite{carla}. The driving vehicle is set up with three front facing cameras. Each data sample of the dataset consists of an image set $\textit{\textbf{I}} = \langle I_l, I_c, I_r\rangle$ of timestamped images from the left, center, and right cameras respectively, paired with their corresponding expert control, in this case, the steering angle for the car collected from the Carla driving autopilot. The augmentations for non-center cameras are computed as shown in Table \ref{tab:driving-corrections}. 

\begin{table}[!t]
\label{table1}
\caption{out-of-distribution experiment results (ground vehicle)}
\begin{center}
\begin{tabular}{|c|c|c|c|}
\hline
\textbf{Training Setup} & \textbf{Interventions}& \textbf{Autonomy (\%)}& \makecell{\textbf{Cross Track} \\ \textbf{Error (m)}} \\
\hline
Center Camera & 8 & 80.24 & 4.00 \\
\hline
\makecell{Equiv. Augment. \\Out-Of-Dist\\Training} & 6 & 87.21 & 2.66 \\
\hline
\makecell{Equiv. Augment. \\In-Dist\\Training} & 5 & 88.01 & 2.61 \\
\hline 
All Cameras & 4 & 89.97 & 2.25 \\
\hline
\end{tabular}
\label{tab:eqfrozen-ground}
\end{center} 
\vspace*{-0.3cm}
\end{table}

\begin{table}[!t]
\caption{Augmentation label calculations (ground vehicle)}
\begin{center}
\vspace*{-0.1cm}
\begin{tabular}{|c|c|c|c|}
\hline
\textbf{Training Camera} & \textbf{Steering Angle} (normalized  $[-1,1]$) \\
\hline
Left Camera & $steering_{cen}+0.05$ \\
\hline
Right Camera & $steering_{cen}-0.05$ \\
\hline
\end{tabular}
\label{tab:driving-corrections}
\end{center}
\vspace*{-0.6cm}
\end{table}


\textbf{Results.} Again we can see in Fig. \ref{fig:cte_drone} (right) that the driving policy trained using equivariant augmentations outperforms the policy trained with just the center camera directly, and that this benefit slowly subsides with more DAgger iterations.

\textbf{Out-Of-Distribution Experiments.} 
As per the flying vehicle experiments, we perform an additional experiment to test generalization of the learned features to novel environments. The experiments so far used a specific path in 'Town10' from the Carla simulator \cite{carla} which resembles a densely packed urban environment with lots of city structures. For this experiment however, the equivariance map was trained using a dataset collected on a highway surrounded by green hills and no other city structures. Yet our results as shown in Table \ref{tab:eqfrozen-ground} illustrate the significant performance improvement gained by our generated augmentations over the policy trained without them.

\begin{figure}[ht]
\centering
\hspace*{-0.5cm}\includegraphics[width=3.5in]{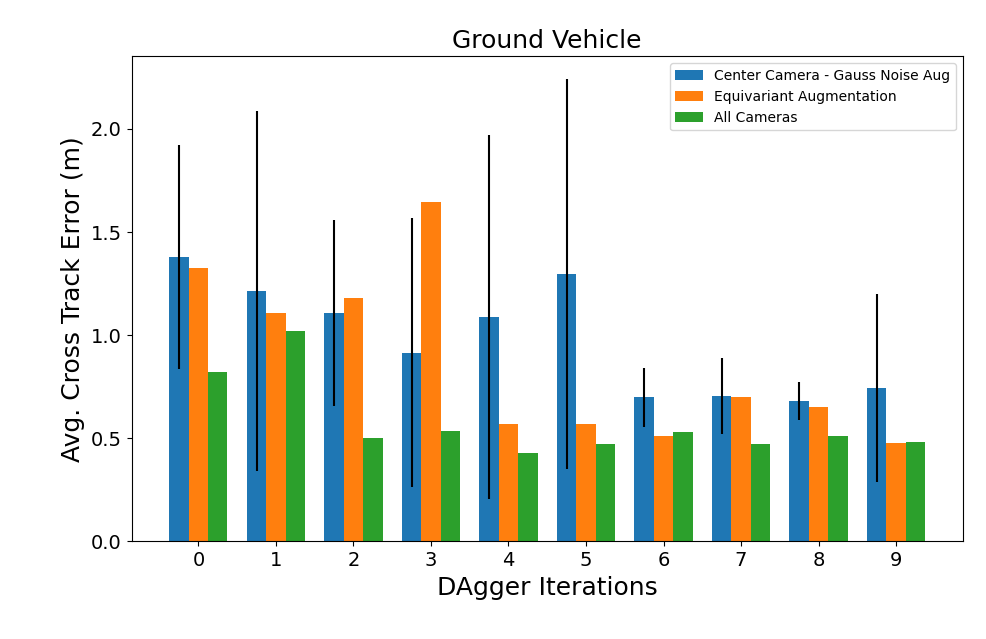}
\vspace*{-0.6cm}
\caption{Average cross track error vs DAgger iterations with center camera network trained with gaussian noise augmentations on the input images.}
\label{fig:gaussnoise}
\vspace{-0.3cm}
\end{figure}

\textbf{Gaussian Noise Augmentation Experiments.} 
\label{noisexp} 
We also compare our method to simply adding noise augmentations to the camera images. The noise augmentations consist of zero mean gaussian noise with varying standard deviations \mbox{$\sigma = \{0.01, 0.05, 0.1, 0.2, 0.3\}$.} While noise augmentations do improve navigation performance, out method still performs better in most cases as shown in Fig.~\ref{fig:gaussnoise}.





\subsection{Terrestrial Robot Experiments} \label{huskyexp}

\begin{figure}[h]
\centering
\includegraphics[width=\columnwidth]{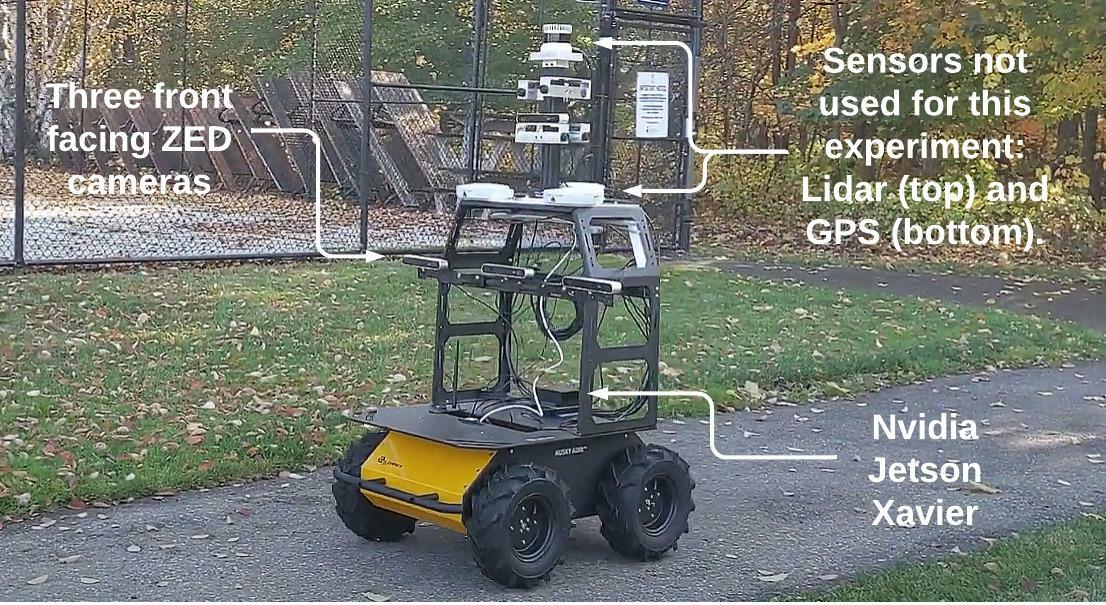}
\vspace{-0.3cm}
\caption{Clearpath Husky used for ground robot experiments with three front facing cameras mounted in the front.}
\label{fig:huskysetup}
\end{figure}

\textbf{Experimental Setup.} 
\label{husksetup}
These experiments were performed using a Clearpath Husky robot equipped with three front facing cameras. The Husky setup is similar to the driving vehicle in Carla simulation as described in Sec.~\ref{carExp}. Three ZED cameras (used as monocular cameras) are mounted on the front, each 25cm apart. All of the computation was performed on the NVIDIA Jetson Xavier mounted on the bottom. The robot can be seen in Fig.~\ref{fig:huskysetup}.



\begin{figure}[!t]
\centering
\includegraphics[width=3.3in]{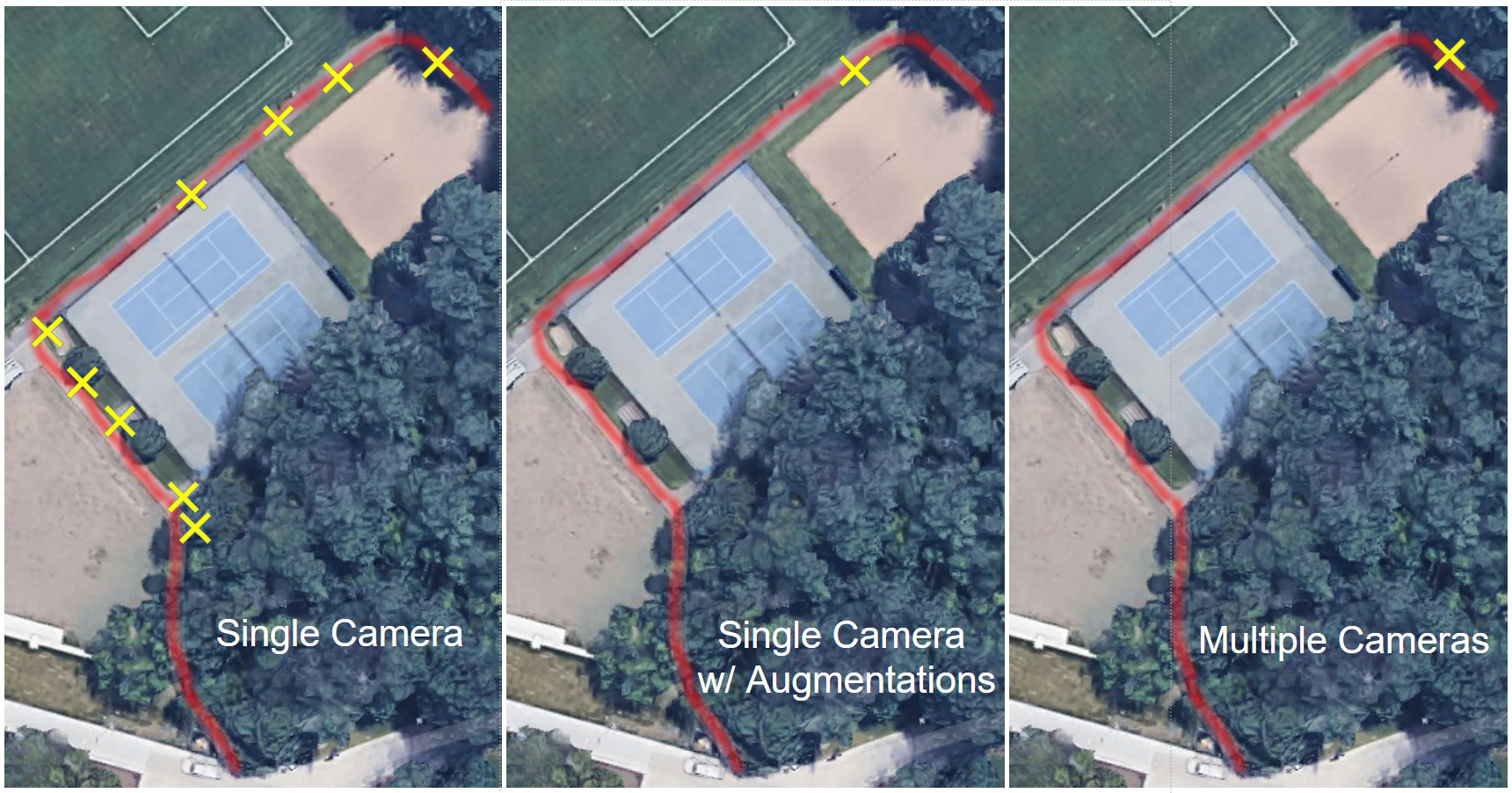}
\caption{A bird's eye view of the testing path for the Husky experiment. Shows the interventions needed while navigating for each policy.}
\label{fig:huskypath}
\end{figure}

\noindent Fig. \ref{fig:huskypath} shows the overhead view of the test path used for the experiment. The path is highlighted in red. Three different models are trained to drive the robot along the path---a model trained using data from a single camera, a model trained using data from a single camera with equivariant augmentations, and a model trained using data from all three cameras.

\textbf{Results.} The model trained using only center camera data performs worse than the others, steering off the path 9 times. The model trained on data generated using equivariant maps as well as the model trained using all three cameras both perform significantly better, steering off the path only once. Fig. \ref{fig:huskypath} shows a snapshot of the entire test run for three setups respectively. The locations where an intervention occurred are denoted by a yellow cross.\footnote{The video from this experiment can be found here: \href{https://youtu.be/5g4Kg3-YWvA}{https://youtu.be/5g4Kg3-YWvA}}

\subsection{Learnt vs Non-Learnt Equivariance Maps}

\begin{figure}[!t]
\centering
\includegraphics[width=3.5in]{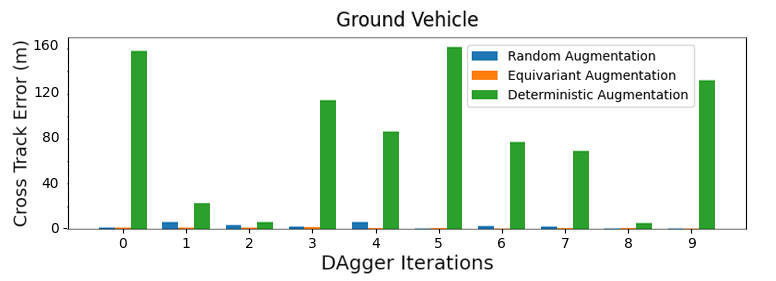}
\caption{Cross track error(m) for driving policy trained using a deterministic map and a learned map enforcing equivariant features.}
\label{fig:detmap_cte}
\vspace{-0.6cm}
\end{figure}

In each of the experiments thus far, we modeled the equivariance maps using a neural network that is optimized using expert data. We investigate the necessity of learning these maps by comparing them to a random map that does not enforce any equivariance relation, and a fixed deterministic map (a linear translation in embedding space). Our results as shown in Fig. \ref{fig:detmap_cte} indicate that both alternative techniques fall short of achieving the performance improvements gained using an equivariance map.

\section{CONCLUSION}

In this paper, we investigate the use of equivariant maps for visual navigation by mobile robots. By training a neural network model that learns image features which are equivariant, we can predict the latent representations of images from viewpoints nearby to the one observed. Our results indicate that by augmenting the training dataset with these representations, one can significantly improve navigation performance in a variety of settings as demonstrated by our simulation experiments involving flying and ground vehicles.
Additionally, through our ground robot experiments over a 500m path, we showed that the benefits of our method also transfer over to real world settings.





\bibliography{bibliography/references.bib}{}

\begin{thebibliography}{10}
\providecommand{\url}[1]{#1}
\csname url@samestyle\endcsname
\providecommand{\newblock}{\relax}
\providecommand{\bibinfo}[2]{#2}
\providecommand{\BIBentrySTDinterwordspacing}{\spaceskip=0pt\relax}
\providecommand{\BIBentryALTinterwordstretchfactor}{4}
\providecommand{\BIBentryALTinterwordspacing}{\spaceskip=\fontdimen2\font plus
\BIBentryALTinterwordstretchfactor\fontdimen3\font minus
  \fontdimen4\font\relax}
\providecommand{\BIBforeignlanguage}[2]{{%
\expandafter\ifx\csname l@#1\endcsname\relax
\typeout{** WARNING: IEEEtran.bst: No hyphenation pattern has been}%
\typeout{** loaded for the language `#1'. Using the pattern for}%
\typeout{** the default language instead.}%
\else
\language=\csname l@#1\endcsname
\fi
#2}}
\providecommand{\BIBdecl}{\relax}
\BIBdecl

\bibitem{pmlr-v97-cohen19c}
J.~Cohen, E.~Rosenfeld, and Z.~Kolter, ``Certified adversarial robustness via
  randomized smoothing,'' in \emph{Proceedings of the 36th International
  Conference on Machine Learning}, ser. Proceedings of Machine Learning
  Research, K.~Chaudhuri and R.~Salakhutdinov, Eds., vol.~97.\hskip 1em plus
  0.5em minus 0.4em\relax PMLR, 09--15 Jun 2019, pp. 1310--1320.

\bibitem{Jayaraman_2015_ICCV}
D.~Jayaraman and K.~Grauman, ``Learning image representations tied to
  ego-motion,'' in \emph{Proceedings of the IEEE International Conference on
  Computer Vision (ICCV)}, December 2015.

\bibitem{lenc2015understanding}
K.~Lenc and A.~Vedaldi, ``Understanding image representations by measuring
  their equivariance and equivalence,'' 2015.

\bibitem{equivariant_rotations}
U.~{Schmidt} and S.~{Roth}, ``Learning rotation-aware features: From invariant
  priors to equivariant descriptors,'' in \emph{2012 IEEE Conference on
  Computer Vision and Pattern Recognition}, 2012, pp. 2050--2057.

\bibitem{Zhao2019QuaternionEC}
Y.~Zhao, T.~Birdal, J.~E. Lenssen, E.~Menegatti, L.~Guibas, and F.~Tombari,
  ``Quaternion equivariant capsule networks for 3d point clouds,''
  \emph{ArXiv}, vol. abs/1912.12098, 2019.

\bibitem{lifd_survey}
\BIBentryALTinterwordspacing
T.~Tuytelaars and K.~Mikolajczyk, ``Local invariant feature detectors: A
  survey,'' \emph{Found. Trends. Comput. Graph. Vis.}, vol.~3, no.~3, p.
  177–280, Jul. 2008. [Online]. Available:
  \url{https://doi.org/10.1561/0600000017}
\BIBentrySTDinterwordspacing

\bibitem{dbn_equivariant}
J.~J. Kivinen and C.~K.~I. Williams, ``Transformation equivariant boltzmann
  machines,'' in \emph{Artificial Neural Networks and Machine Learning -- ICANN
  2011}, T.~Honkela, W.~Duch, M.~Girolami, and S.~Kaski, Eds.\hskip 1em plus
  0.5em minus 0.4em\relax Berlin, Heidelberg: Springer Berlin Heidelberg, 2011,
  pp. 1--9.

\bibitem{Bilovs2020EquivariantNF}
M.~Bilovs and S.~Gunnemann, ``Equivariant normalizing flows for point processes
  and sets,'' 2020.

\bibitem{Rezende2019EquivariantHF}
D.~J. Rezende, S.~Racani{\`e}re, I.~Higgins, and P.~T{\'o}th, ``Equivariant
  hamiltonian flows,'' \emph{ArXiv}, vol. abs/1909.13739, 2019.

\bibitem{Weiler20183DSC}
M.~Weiler, M.~Geiger, M.~Welling, W.~Boomsma, and T.~Cohen, ``3d steerable
  cnns: Learning rotationally equivariant features in volumetric data,'' in
  \emph{NeurIPS}, 2018.

\bibitem{Spezialetti2019LearningAE}
R.~Spezialetti, S.~Salti, and L.~Stefano, ``Learning an effective equivariant
  3d descriptor without supervision,'' \emph{2019 IEEE/CVF International
  Conference on Computer Vision (ICCV)}, pp. 6400--6409, 2019.

\bibitem{Fuchs2020SE3Transformers3R}
F.~Fuchs, D.~E. Worrall, V.~Fischer, and M.~Welling, ``Se(3)-transformers: 3d
  roto-translation equivariant attention networks,'' \emph{ArXiv}, vol.
  abs/2006.10503, 2020.

\bibitem{Wang2020EquivariantMF}
R.~Wang, M.~Albooyeh, and S.~Ravanbakhsh, ``Equivariant maps for hierarchical
  structures,'' \emph{ArXiv}, vol. abs/2006.03627, 2020.

\bibitem{cubenet}
D.~Worrall and G.~Brostow, ``Cubenet: Equivariance to 3d rotation and
  translation,'' in \emph{Computer Vision -- ECCV 2018}, V.~Ferrari, M.~Hebert,
  C.~Sminchisescu, and Y.~Weiss, Eds.\hskip 1em plus 0.5em minus 0.4em\relax
  Cham: Springer International Publishing, 2018, pp. 585--602.

\bibitem{worrall2017harmonic}
D.~E. Worrall, S.~J. Garbin, D.~Turmukhambetov, and G.~J. Brostow, ``Harmonic
  networks: Deep translation and rotation equivariance,'' 2017.

\bibitem{cohen2019gauge}
T.~S. Cohen, M.~Weiler, B.~Kicanaoglu, and M.~Welling, ``Gauge equivariant
  convolutional networks and the icosahedral cnn,'' 2019.

\bibitem{Tai2019EquivariantTN}
K.~S. Tai, P.~Bailis, and G.~Valiant, ``Equivariant transformer networks,''
  \emph{ArXiv}, vol. abs/1901.11399, 2019.

\bibitem{Cohen2019AGT}
T.~Cohen, M.~Geiger, and M.~Weiler, ``A general theory of equivariant cnns on
  homogeneous spaces,'' \emph{ArXiv}, vol. abs/1811.02017, 2019.

\bibitem{steerable_filters_equivariant_cnn}
M.~{Weiler}, F.~A. {Hamprecht}, and M.~{Storath}, ``Learning steerable filters
  for rotation equivariant cnns,'' in \emph{2018 IEEE/CVF Conference on
  Computer Vision and Pattern Recognition}, 2018, pp. 849--858.

\bibitem{emvn}
C.~{Esteves}, Y.~{Xu}, C.~{Allec-Blanchette}, and K.~{Daniilidis},
  ``Equivariant multi-view networks,'' in \emph{2019 IEEE/CVF International
  Conference on Computer Vision (ICCV)}, 2019, pp. 1568--1577.

\bibitem{multiviews}
A.~{Kanezaki}, Y.~{Matsushita}, and Y.~{Nishida}, ``Rotationnet: Joint object
  categorization and pose estimation using multiviews from unsupervised
  viewpoints,'' in \emph{2018 IEEE/CVF Conference on Computer Vision and
  Pattern Recognition}, 2018, pp. 5010--5019.

\bibitem{gvcnn}
Y.~{Feng}, Z.~{Zhang}, X.~{Zhao}, R.~{Ji}, and Y.~{Gao}, ``Gvcnn: Group-view
  convolutional neural networks for 3d shape recognition,'' in \emph{2018
  IEEE/CVF Conference on Computer Vision and Pattern Recognition}, 2018, pp.
  264--272.

\bibitem{qi2016volumetric}
C.~R. Qi, H.~Su, M.~Nie{\ss}ner, A.~Dai, M.~Yan, and L.~Guibas, ``Volumetric
  and multi-view cnns for object classification on 3d data,'' in \emph{Proc.
  Computer Vision and Pattern Recognition (CVPR), IEEE}, 2016.

\bibitem{you2018pvnet}
H.~You, Y.~Feng, R.~Ji, and Y.~Gao, ``Pvnet: A joint convolutional network of
  point cloud and multi-view for 3d shape recognition,'' 2018.

\bibitem{Pol2020PlannableAT}
E.~van~der Pol, T.~Kipf, F.~A. Oliehoek, and M.~Welling, ``Plannable
  approximations to mdp homomorphisms: Equivariance under actions,'' in
  \emph{AAMAS}, 2020.

\bibitem{ratner2017learning}
A.~J. Ratner, H.~R. Ehrenberg, Z.~Hussain, J.~Dunnmon, and C.~Ré, ``Learning
  to compose domain-specific transformations for data augmentation,'' 2017.

\bibitem{zhang2019adversarial}
X.~Zhang, Q.~Wang, J.~Zhang, and Z.~Zhong, ``Adversarial autoaugment,'' 2019.

\bibitem{dao2019kernel}
T.~Dao, A.~Gu, A.~J. Ratner, V.~Smith, C.~D. Sa, and C.~Ré, ``A kernel theory
  of modern data augmentation,'' 2019.

\bibitem{raileanu2020automatic}
R.~Raileanu, M.~Goldstein, D.~Yarats, I.~Kostrikov, and R.~Fergus, ``Automatic
  data augmentation for generalization in deep reinforcement learning,'' 2020.

\bibitem{ravi_unsupervised_depth}
R.~Garg, V.~K. B.G., G.~Carneiro, and I.~Reid, ``Unsupervised cnn for single
  view depth estimation: Geometry to the rescue,'' in \emph{Computer Vision --
  ECCV 2016}, B.~Leibe, J.~Matas, N.~Sebe, and M.~Welling, Eds.\hskip 1em plus
  0.5em minus 0.4em\relax Cham: Springer International Publishing, 2016, pp.
  740--756.

\bibitem{Loquercio_2018}
A.~Loquercio, A.~I. Maqueda, C.~R.~D. Blanco, and D.~Scaramuzza, ``Dronet:
  Learning to fly by driving,'' \emph{{IEEE} Robotics and Automation Letters},
  2018.

\bibitem{lecun_obstacle_avoidance}
U.~Muller, J.~Ben, E.~Cosatto, B.~Flepp, and Y.~L. Cun, ``Off-road obstacle
  avoidance through end-to-end learning,'' in \emph{Advances in Neural
  Information Processing Systems 18}.\hskip 1em plus 0.5em minus 0.4em\relax
  MIT Press, 2006, pp. 739--746.

\bibitem{pomerleau_thesis}
D.~Pomerleau, ``Neural network perception for mobile robot guidance,'' Ph.D.
  dissertation, Carnegie Mellon University, 1993.

\bibitem{alvinn}
D.~A. Pomerleau, ``Alvinn: An autonomous land vehicle in a neural network,'' in
  \emph{Advances in Neural Information Processing Systems 1}, D.~S. Touretzky,
  Ed., 1989, pp. 305--313.

\bibitem{DAgger_ross}
S.~Ross, G.~J. Gordon, and D.~Bagnell, ``A reduction of imitation learning and
  structured prediction to no-regret online learning,'' in \emph{International
  Conference on Artificial Intelligence and Statistics, {AISTATS}}, 2011, pp.
  627--635.

\bibitem{trail_following}
A.~Giusti, J.~Guzzi, D.~C. Cireşan, F.~L. He, J.~P. Rodriguez, F.~Fontana,
  M.~Faessler, C.~Forster, J.~Schmidhuber, G.~D. Caro, D.~Scaramuzza, and L.~M.
  Gambardella, ``A machine learning approach to visual perception of forest
  trails for mobile robots,'' \emph{IEEE Robotics and Automation Letters},
  vol.~1, no.~2, pp. 661--667, July 2016.

\bibitem{nvidia_driving}
\BIBentryALTinterwordspacing
M.~Bojarski, D.~D. Testa, D.~Dworakowski, B.~Firner, B.~Flepp, P.~Goyal, L.~D.
  Jackel, M.~Monfort, U.~Muller, J.~Zhang, X.~Zhang, J.~Zhao, and K.~Zieba,
  ``End to end learning for self-driving cars,'' \emph{CoRR}, vol.
  abs/1604.07316, 2016. [Online]. Available:
  \url{http://arxiv.org/abs/1604.07316}
\BIBentrySTDinterwordspacing

\bibitem{codevilla_2017}
F.~Codevilla, M.~M{\"{u}}ller, A.~Dosovitskiy, A.~L{\'{o}}pez, and V.~Koltun,
  ``End-to-end driving via conditional imitation learning,'' \emph{CoRR}, vol.
  abs/1710.02410, 2017.

\bibitem{dart_laskey}
M.~Laskey, J.~Lee, R.~Fox, A.~Dragan, and K.~Goldberg, ``Dart: Noise injection
  for robust imitation learning,'' in \emph{Proceedings of the 1st Annual
  Conference on Robot Learning}, ser. Proceedings of Machine Learning Research,
  vol.~78.\hskip 1em plus 0.5em minus 0.4em\relax PMLR, 13--15 Nov 2017, pp.
  143--156.

\bibitem{adam}
\BIBentryALTinterwordspacing
D.~P. Kingma and J.~Ba, ``Adam: {A} method for stochastic optimization,'' in
  \emph{3rd International Conference on Learning Representations, {ICLR} 2015,
  San Diego, CA, USA, May 7-9, 2015, Conference Track Proceedings}, Y.~Bengio
  and Y.~LeCun, Eds., 2015. [Online]. Available:
  \url{http://arxiv.org/abs/1412.6980}
\BIBentrySTDinterwordspacing

\bibitem{DAgger}
S.~Ross, G.~Gordon, and D.~Bagnell, ``A reduction of imitation learning and
  structured prediction to no-regret online learning,'' in \emph{Proceedings of
  the fourteenth international conference on artificial intelligence and
  statistics}, 2011, pp. 627--635.

\bibitem{airsim2017fsr}
\BIBentryALTinterwordspacing
S.~Shah, D.~Dey, C.~Lovett, and A.~Kapoor, ``Airsim: High-fidelity visual and
  physical simulation for autonomous vehicles,'' in \emph{Field and Service
  Robotics}, 2017. [Online]. Available: \url{https://arxiv.org/abs/1705.05065}
\BIBentrySTDinterwordspacing

\bibitem{droneracing}
R.~Madaan, N.~Gyde, S.~Vemprala, M.~Brown, K.~Nagami, T.~Taubner,
  E.~Cristofalo, D.~Scaramuzza, M.~Schwager, and A.~Kapoor, ``Airsim drone
  racing lab,'' \emph{arXiv preprint arXiv:2003.05654}, 2020.

\bibitem{carla}
A.~Dosovitskiy, G.~Ros, F.~Codevilla, A.~Lopez, and V.~Koltun, ``{CARLA}: {An}
  open urban driving simulator,'' in \emph{Proceedings of the 1st Annual
  Conference on Robot Learning}, 2017, pp. 1--16.

\end{thebibliography}
\bibliographystyle{IEEEtran}

\end{document}